# HyperFaceNet: A Hyperspectral Face Recognition Method Based on Deep Fusion

Zhicheng Cao [a], Xi Cen [a] and Liaojun Pang [a,*]

[a] School of Life Science and Technology, Xidian University, Xi'an 710071, China.

**Abstract**
Face recognition has already been well studied under the visible light and the infrared, in both intra-spectral and cross-spectral cases. However, how to fuse different light bands, i.e., hyperspectral face recognition, is still an open research problem, which has the advantages of richer information retaining and all-weather functionality over single band face recognition. Among the very few works for hyperspectral face recognition, traditional non-deep learning techniques are largely used. Thus, we in this paper bring deep learning into the topic of hyperspectral face recognition, and propose a new fusion model (named HyperFaceNet) especially for hyperspectral faces. The proposed fusion model is characterized by residual dense learning, a feedback style encoder and a recognition-oriented loss function. During the experiments, our method is proved to be of higher recognition rates than face recognition using either visible light or the infrared. Moreover, our fusion model is shown to be superior to other general-purposed image fusion methods including state-of-the-arts, in terms of both image quality and recognition performance.

*Send correspondence to Liaojun Pang: ljpang@mail.xidian.edu.cn

## I. Introduction

In the past few decades, many scholars have studied face recognition based on visible light. But, due to the change of illumination and other environmental settings, the performance of visible light-based face recognition method can be easily affected. Such issue causes serious problems in many real-world applications where the lighting condition is most likely uncontrolled, resulting in failures of the face recognition technology.

The invariance of infrared (IR) imaging to the change of light is helpful for alleviating this issue and thus improving the recognition performance [1]. Besides, IR is able to see through the darkness and continues to work under harsh atmospheric conditions such as rain, snow and fog. Thermal IR images are advantageous for the ability of capturing the patterns associated with heat emission of an object, such that it can extract additional anatomical structures and even can distinguish between identical twins. Therefore, a great amount of interests has been diverted to the research of infrared-based face recognition [2][3][4][5].

However, compared with the visible image, the resolution of the infrared image is usually lower, and the texture features of the image are not clear enough [6]. As a result, there have been researchers who find hyperspectral face recognition as an alternative, i.e., to fuse the visible light image with the IR image. Because it utilizes information from both light wavebands (i.e., visible light and IR), hyperspectral face recognition usually leads to an improved recognition rate compared with any waveband. For instance, Gyaourova *et al.* first introduced the method of image fusion into face recognition, fused infrared face image and visible face image in the wavelet domain, used genetic algorithm (GAs) to select fusion information, and finally obtained a fused face image [7]. The experiment proves that this method can effectively improve the accuracy of face recognition.

The problem, however, is that all the existing algorithms of hyperspectral face recognition adopt traditional hand-crafted operators which either fail to perform well enough in practice or have poor generality in different cases. Hence, in view of this, we in this paper bring deep learning into the topic of hyperspectral face recognition. To our best knowledge, our work is the first of such efforts. We propose a novel fusion model to fuse up IR and visible faces and name it HyperFaceNet. The model is designed especially for fusion of hyperspectral faces, characterized by residual dense learning, a feedback style encoder and a recognition-oriented loss function.

The rest of the paper is organized as follow: In Section II, we briefly review related works. In Section III, the proposed hyperspectral face image fusion method based on deep learning is introduced in details. The experimental results are shown in Section IV. The conclusion of our paper with discussion are presented in Section V.

## II. Related Work

There exist a number of works in the literature that conduct hyperspectral face recognition by fusing the visible light and the IR facial images. However, hyperspectral face recognition using deep learning has not been studied so far. We review a few existing publications on the topic.

In 2004, Gyaourova and Bebis [7] first introduced the method of image fusion into face recognition. They decomposed infrared face image and visible face image in the wavelet domain, used genetic algorithm (GAs) to select fusion information, and finally obtained face fusion image. The experiment of face fusion proves that this method can effectively improve the accuracy of face recognition.

In 2008, Singh R et al. [8] used the Discrete wavelet transform (DWT) method combined with Gabor transform to extract the amplitude and phase features of the fused image, and then used the adaptive SVM method to merge the extracted features, obtaining the recognition effect with an error rate of 2.86% and the highest recognition accuracy at that time.

Since then, more fusion method is applied to the face image fusion [9-14], as proposed by Chen et al., based on integral wavelet frame transform (Nonseparable wavelet frame transform, NWFT) and ICA the fusion recognition method of combining the registered NWFT light image and infrared image fusion, the ICA to extract facial image independent characteristics, Top - match method to classify fusion face [15].

Ma et al. designed a matching fraction fusion method based on two-dimensional linear discriminant analysis, and applied Euclidean distance corresponding similarity to perform similarity fusion [16].

In their review in 2018, Jain et al. made a detailed comparison of the fusion effect and recognition effect of existing traditional face image fusion algorithms, and pointed out that traditional methods have problems such as manual design fusion rules, fusion parameter selection relying on experience, and fusion image artifacts [17].

With the development of the deep learning technology, the latest research works have introduced deep learning into the general problem of image fusion. The deep learning based fusion methods can effectively avoid issues such as manual design of fusion rules and selection of fusion parameters and have better fusion effect than traditional fusion algorithms [18-21]. Therefore, we in this paper propose a hyperspectral face fusion network based on deep learning to achieve more accurate face recognition.

The above results have shown that the face image fusion of different spectra can effectively improve the face recognition accuracy. However, according to our survey, there is still a lack of deep fusion method on the topic of human face image fusion. The particular problem of fusing different spectrum of human face image need special treatment. Whether an image fusion method based on deep learning can ultimately improve the recognition performance effect remains to be studied and explored.

In this paper, our main contributions are:

(1) We bring the technology of deep learning to the topic of hyperspectral face recognition to tackle shortcomings of traditional methods. According to our best knowledge, we are the first to deal with the hyperspectral face recognition problem with a deep learning approach.

(2) A hyperspectral face image fusion model based on deep learning was proposed (named HyperFaceNet), which is featured by a Siamese encoder-decoder structure and includes an extra module of pre-fusion.

(3) Multiple Residual Dense Blocks with local and global learning are introduced to enhance the feature extraction ability of the network. A feedback technology is involved in the decoder to improve the performance.

(4) We propose a composite loss function by considering critical components such as structural loss, pixel loss and facial detail loss. Such a composite loss is face recognition-oriented and beneficial for the final recognition accuracy.

(5) We train the face recognition model based on convolutional neural network and the triad loss function by using the face data set after the hyperspectral fusion, involving transfer learning. We conducted experiments on three multi-spectral face data sets of QFIRE and CASIA, and both the fusion image quality and recognition accuracy are higher than other existing algorithms.

## III. Proposed Method

In this section, the proposed deep learning-based hyperspectral face image fusion method is explained in details. It is mainly described in three aspects: the network architecture, the training process and the fusion strategy.

### A. Network Architecture

The input infrared and visible face images (gray level images) are denoted as $I_i$ and $I_v$, respectively. We assume that the input images have been registered. Our network architecture comprises four parts: the pre-fusion layer, the encoder, the fusion layer and the decoder. The overall architecture of the proposed network is shown in Fig.1.

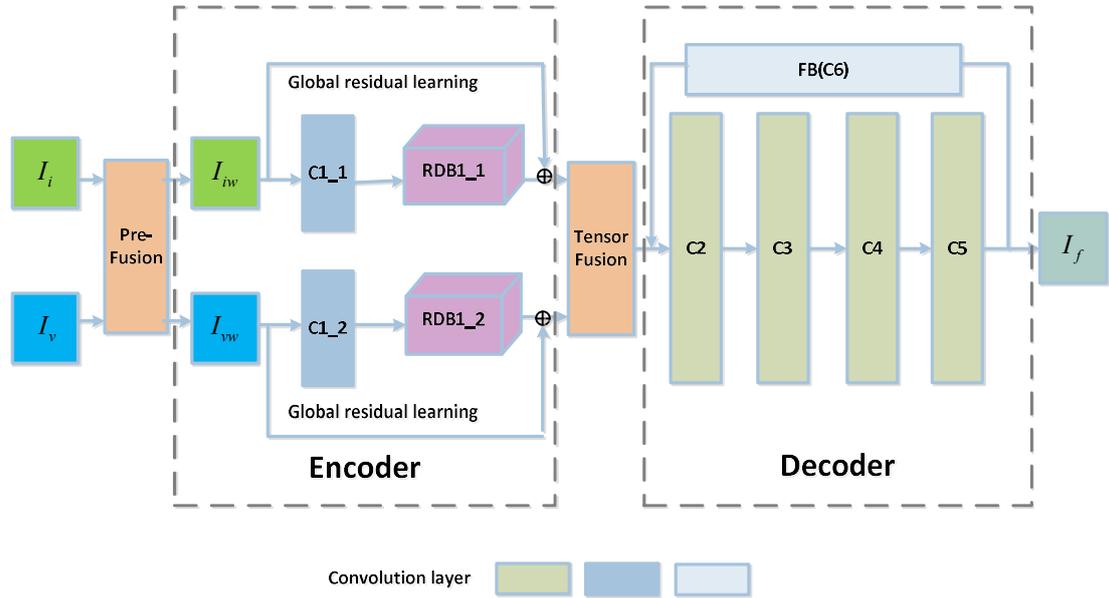

Fig.1 The overall architecture of the proposed network (HyperFaceNet)

As shown in Fig.1, the pre-fusion layer is the weighted fusion of infrared image and visible image, so as to obtain a group of simple fusion images with a large weight of visible image ($I_{iw}$) and infrared image ($I_{vw}$) respectively. Then, $I_{iw}$ and $I_{vw}$ are input to the following encoder.

The encoder is a Siamese architecture network which has two channels (C1_1 and RDB1_1 for channel1, C1_2 and RDB1_2 for channel2). The first convolution layer (C1_1 and C1_2) is used to extract rough features and the residual dense block (RDB1_1 and RDB1_2) (as shown in Fig.2 ) contains five convolution layers (the first three convolution layers are connected in a dense way and the output is splicing at multiple scales[18]. Then the number of feature images is adjusted through the convolutional layer, and the local residual skip learning is connected at last.) are used to make full use of the information of all local layers to improve the network representation. In addition, the entire encoder is connected by global residual skip learning. The weights of encoder channels are tied, C1_1 and C1_2 (RDB1_1 and RDB1_2) share the same weights. We choose one fusion strategy in fusion layer and it will be introduced in Section III-C.

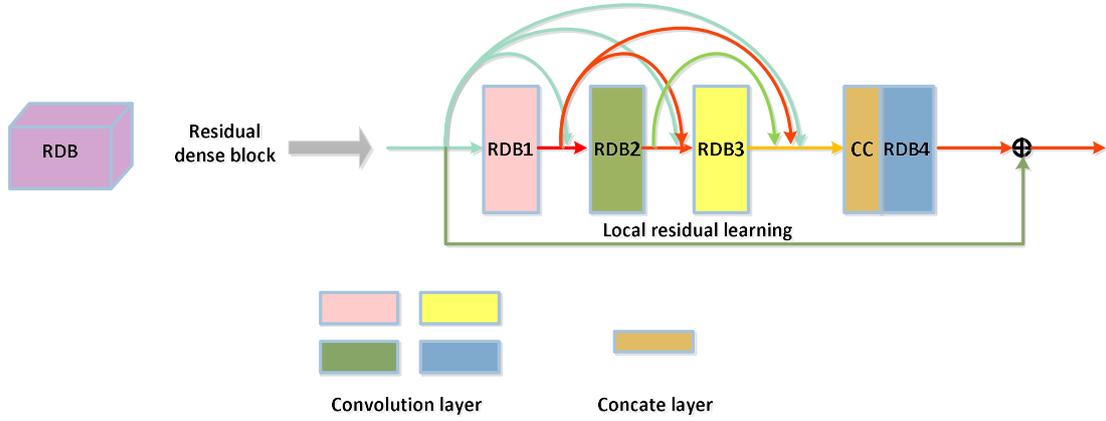

Fig.2 The architecture of residual dense blocks.

The decoder contains five convolution layers. The last convolution layer (C6) is used to for feedback connection. The use of feedback linkage can correct the input with the output image to obtain a higher quality reconstructed image. The number of feedback connections used in this paper is 4, and the specific structure is shown in Fig.3.

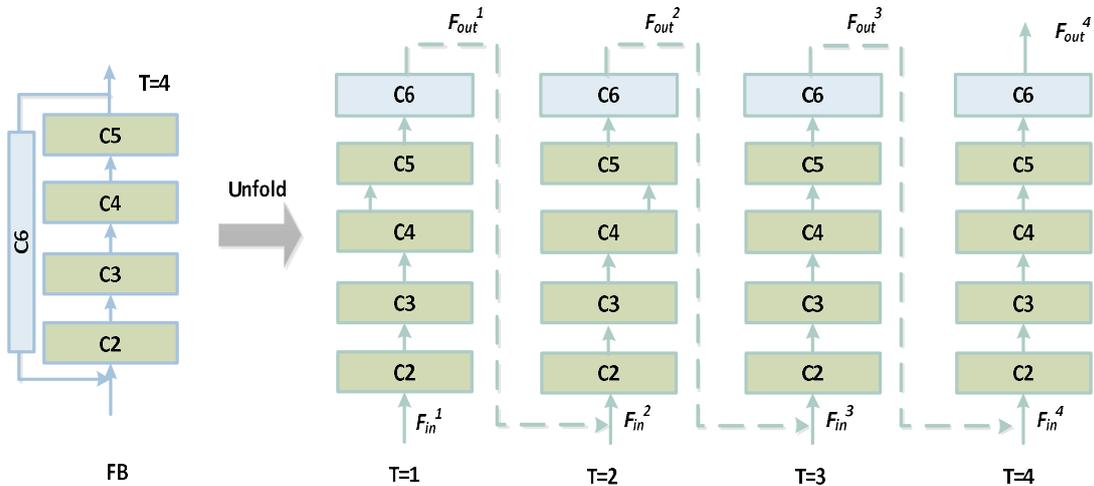

Fig.3 The feedback connections unfolded.

## B. Composite Loss

In training process, we only consider encoder and decoder network. We train encoder and decoder to reconstruct the input image. In order to reconstruct the input fusion image more precisely, we designed a composite loss function L to train our encoder and decoder,

$$L = \lambda L_{ssim} + L_p + \gamma L_{ag} \quad (1)$$

which is a weighted combination of pixel loss $L_p$, structural similarity (SSIM) loss $L_{ssim}$ with the weight $\lambda$ and the average gradient loss $L_{ag}$ with the weight $\gamma$.

The pixel loss $L_p$ is calculated as,

$$L_p = \|O - I\|_2 \quad (2)$$

where $O$ and $I$ indicate the output and input image, respectively. It is the Euclidean distance between the output $O$ and the input $I$.

The structural similarity loss $L_{ssim}$ is obtained by,

$$L_{ssim} = 1 - SSIM(O, I) \quad (3)$$

where SSIM(·) represents the structural similarity operation[22] and it denotes the structural similarity of two images.

The average gradient loss $L_{ag}$ is defined by,

$$L_{ag} = \frac{1}{M \times N} \sum_{i=1}^{M} \sum_{j=1}^{N} \sqrt{\frac{\left(\frac{\partial O}{\partial x}\right)^2 + \left(\frac{\partial O}{\partial y}\right)^2}{2}} \quad (4)$$

where $M \times N$ represents the size of the image, $\frac{\partial O}{\partial x}$ and $\frac{\partial O}{\partial y}$ denote the output image gradients in horizontal and vertical directions, respectively.

We trained out network using the infrared images and visible images of CASIA and QFIRE (two IR face datasets) as the input, all of which are resized to 256×256 and the RGB images are transformed to gray ones. The learning rate is set as $1 \times 10^{-4}$. The batch size and epochs are 32 and 200, respectively. Our method is implemented on a computer with GTX 1080Ti and 8GB RAM.

## C. Fusion Strategy
### 1) Pre-fusion layer

The purpose of pre-fusion layer is to make the network make full use of the information of different spectral images in the training process, to avoid using a single spectral image for training, resulting in the fusion image to over-display the characteristics of a spectrum, thus affecting the face recognition accuracy.

The output $I_{iw}$ and $I_{vw}$ of the pre-fusion layer can be obtained by,

$$\begin{cases} I_{iw} = a_1 I_i + a_2 I_v \\ I_{vw} = a_2 I_i + a_1 I_v \end{cases} \quad (5)$$

where $a_1 + a_2 = 1$.

*2) Fusion layer*

Once the encoder and decoder networks are trained, in testing process, we used two-stream architecture in encoder and the weight are tied. We choose the addition strategy to combine salient feature maps which are obtain by encoder.

The addition fusion strategy is simply realized by adding up the feature maps of the infrared and the visible light facial images, which is shown in Fig.4.

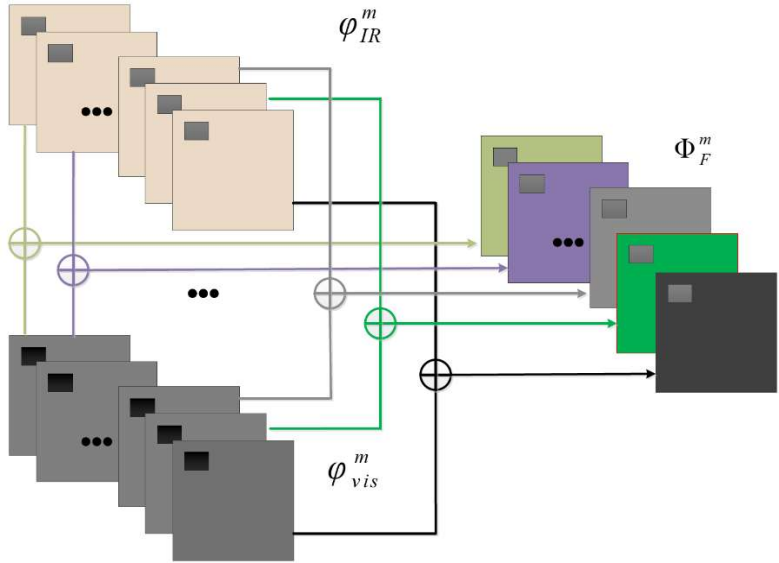

Fig.4 The procedure of addition strategy.

Let $\varphi_1^m$ and $\varphi_2^m$ denote the feature maps which are obtained by the encoder from input images, and $y^m$ denotes the fused feature maps. The addition strategy is formulated by

$$y^m(x, y) = \varphi_1^m(x, y) + \varphi_2^m(x, y) \quad (6)$$

where $(x, y)$ denotes the corresponding position in feature maps and fused feature maps. $y^m$ serves as the input of the following decoder and final fused image will be reconstructed by the decoder.

The specific parameters of the proposed network are summarized and listed in Table I.

Table I. The specific parameters of the proposed network. Conv denotes the convolutional layer; RDB denotes the residual dense lock.

| Module Name | Layer | Kernel Size | Channels (input) | Channels (output) | Activation |
|---|---|---|---|---|---|
| Encoder | Conv(C1) | 3 | 1 | 16 | ReLu |
| | RDB | 3 | 16 | 64 | ReLu |
| Decoder | Conv(C2) | 3 | 64 | 64 | ReLu |
| | Conv(C3) | 3 | 64 | 32 | ReLu |
| | Conv(C4) | 3 | 32 | 16 | ReLu |
| | Conv(C5) | 3 | 16 | 1 | - |
| | FB | 3 | 1 | 64 | - |
| RDB (residual dense block) | Conv(RDB1) | 3 | 16 | 16 | ReLu |
| | Conv(RDB2) | 3 | 32 | 16 | ReLu |
| | Conv(RDB3) | 3 | 48 | 16 | ReLu |
| | Concate(CC) | 3 | 64 | 64 | - |
| | Conv(RDB4) | 1 | 64 | 64 | - |
| FB (Feedback) | Conv(C6) | 3 | 1 | 64 | - |

## IV. Experimental Results and Analysis

In this section, we first introduce the datasets, then use subjective and objective criteria to evaluate the proposed fusion method and compare with existing methods. Finally, we carry out the experiment of hyperspectral face recognition, and demonstrate the superiority of the proposed algorithm from the perspective of recognition accuracy.

*A. Datasets*

We use visible face images and near-infrared face images from two publicly available data sets, which are CASIA NIR-VIS collected by the Chinese Academy of Sciences and QFIRE collected by Clarkson University. Both datasets include face images in IR and the visible light. The proposed network was trained and tested in a ratio of 3:1. A sample of these images is shown in Fig.5.

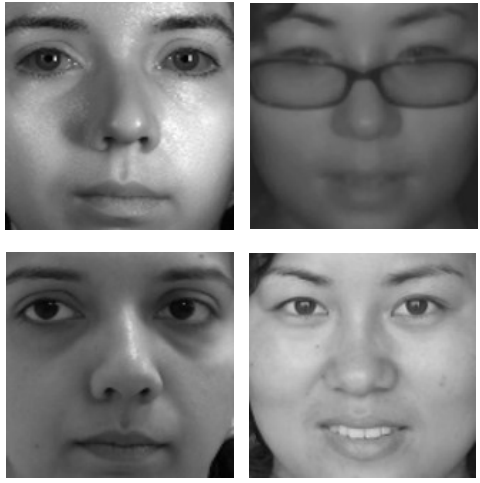

Fig.5. Two pairs of source images from different datasets, from left to right are CASIA and QFIRE. The top row are the infrared face images, and the second row are the visible light face images.

*B. Fusion Quality Evaluation*

We compare the proposed method with several typical fusion methods, including wavelet transform fusion method (WT) [23], cross bilateral filter fusion method (CBF) [24], the joint-spare representation mode (JSR) [25], and the deep learning-based method DenseFuse[26]. In our experiment, the filter size is set as 3×3 for DenseFuse method.

For the purpose of quantitative comparison between our fusion method and other existing algorithms, four quality metrics are utilized. These are entropy (EN) [27], image edge fidelity ($Q_{abf}$) [28], modified structural similarity for no-reference image (SSIM) [29], and the well-known metrics of peak signal-to-noise ratio (PSNR) [30]. The fusion performance improves as the values of all the four metrics increase.

The fused images obtained by the four existing methods and the proposed method use which are shown in Fig.6 and Fig.7. Due to the space limit, we evaluate the relative performance of the fusion methods on two images from two datasets, respectively.

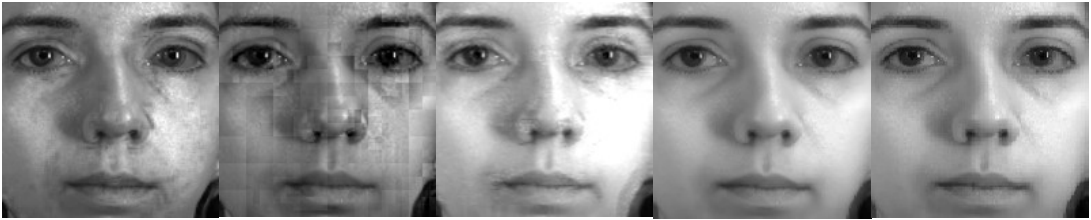

Fig.6 The results of different fusion methods on the CASIA dataset. From left to right are: CBF, WT, JSR, DenseFuse, our proposed method.

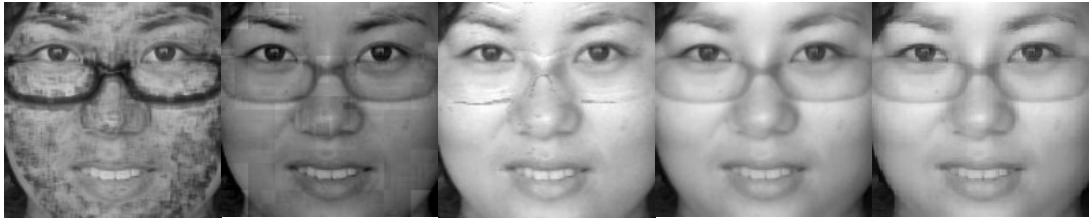

Fig.7 The results of different fusion methods on the QFIRE dataset. From left to right are: CBF, WT, JSR, DenseFuse, our proposed method.

The fused images obtained by CBF and WT demonstrate obvious artificial noise, and the saliency features are not clear, as we can see from Fig.6 and Fig.7. In Fig.6, there is some artificial noise on the fused images when JSR method is utilized to fuse images, and the brow of the face image is not fused well in Fig.7.

Compared with DenseFuse, our proposed method (HyperFaceNet) can obtain a bit clearer fused images, but there is no validate difference between them in human sensitivity. Therefore, we choose several objective metrics to evaluate the fusion performance in the next.

The average values of four metrics for all fused images in two datasets which are obtained by existing methods and the proposed fusion method are shown in Table II.

Table II The values of quality metrics for different fusion methods
on the two datasets, CASIA and QFIRE.

| Datasets | Fusion methods | EN | $Q_{abf}$ | SSIM | PSNR |
|---|---|---|---|---|---|
| CASIA | CBF[24] | 6.7575 | 0.2500 | 0.7001 | 17.32 |
| | WT[23] | 6.5934 | 0.3040 | 0.8238 | 15.87 |
| | JSR[25] | 7.0717 | 0.2990 | 0.8187 | 20.78 |
| | DenseFuse[26] | 7.2001 | 0.5913 | 0.9509 | 23.02 |
| | **Our proposed** | **7.2022** | **0.5913** | **0.9514** | **23.06** |
| QFIRE | CBF[24] | **7.7932** | 0.2005 | 0.5078 | 17.32 |
| | WT[23] | 7.4176 | 0.1904 | 0.5127 | 15.08 |
| | JSR[25] | 7.7064 | 0.2075 | 0.5854 | 19.35 |
| | DenseFuse[26] | 7.6959 | 0.4963 | 0.9224 | **21.35** |
| | **Our proposed** | 7.7522 | **0.5463** | **0.9386** | 21.03 |

The best values for quality metrics are indicated in bold and the second-best values are indicated in blue. As we can see, the proposed method achieves the best value under all the four metrics (i.e., EN, $Q_{abf}$, SSIM and PSNR) on the CASIA dataset while it achieves the best two out of the four metrics on QFIRE (i.e., $Q_{abf}$ and SSIM). In the latter case of CASIA, the other two metrics come at the second best for our method.

Overall, our method yields the best performance in terms of comprehensive image quality. The face that our method achieves the best values in SSIM suggests that our method can preserve more structural details. It also indicates our method can obtain more features and information and introduce less artificial noise from the images to be fused since it achieves the best or the second-best values in EN, $Q_{abf}$ and PSNR.

*C. Recognition Performance Evaluation*

In this part, we will verify the performance of the proposed hyperspectral face fusion method from another perspective. We use the accuracy of face recognition to illustrate the effect of the proposed method on recognition performance and further demonstrate the superiority of the method.

We used the five methods in the previous part to obtain five groups of hyperspectral face fusion data on different data sets, and then used the face recognition network FaceNet and support vector machine (SVM) to identify the obtained data [29]. The recognition results are shown in Table III (We also compared the recognition accuracy with that of a single spectral face).

Table III. Comparison between the recognition rates of different fusion methods on the two datasets, CASIA and QFIRE.

| **Methods** | **Test datasets** | |
| --- | --- | --- |
| | CASIA | QFIRE |
| Visible light | 87.5% | 88.6% |
| Infrared | 81.2% | 81.8% |
| CBF[24] | 70.8% | 70.2% |
| WT[23] | 81.2% | 89.1% |
| JSR[25] | 83.3% | 87.5% |
| DenseFuse[26] | 85.4% | 93.5% |
| **Our proposed** | **89.6%** | **95.7%** |

As can be seen from Table III, compared with other four methods, the proposed method achieves the best recognition effect on different data sets (89.6% on CASIA and 95.7% on QFIRE), and the recognition rate is significantly higher than the result of single-spectrum recognition (i.e., either visible light or infrared alone). These results show that the proposed method can improve the quality of hyperspectral fusion face image and further improve the accuracy of face recognition, which verifies the superiority of the proposed method.

**V. Conclusion**

In this paper, we propose a hyperspectral face recognition model (HyperFaceNet) based on deep learning to alleviate the shortcomings of using a single waveband of light in face recognition. As far as we know, such an effort is among the first in the literature.

The proposed network is characterized by a Siamese encoder-decoder structure and consists of four modules: pre-fusion layer, encoder, fusion layer and decoder. Firstly, the original infrared and visible light face images are utilized to be the input of pre-fusion layer. The purpose of pre-fusion layer is to make full use of the information of multispectral face images during training. Then the features maps are obtained by CNN layers and RDB blocks. After the fusion layer, the feature maps are integrated into one feature map which contains all salient features from the multispectral images. Finally, the fused face image is reconstructed by a feedback-style decoder.

We use both subjective and objective image quality metrics to evaluate our fusion method, and the face recognition rate is further used to justify the model. All the experimental results show that the proposed method exhibits state-of-the-art fusion performance and recognition rates.

**Acknowledgment**

This research is funded by the National Natural Science Foundation of China under Grant 61906149, the Fundamental Research Funds for the Central Universities under Grant JB181206 and the National Cryptography Development Fund under Grant MMJJ20170208.